\documentclass[10pt,journal,compsoc]{IEEEtran}
\usepackage{times}
\usepackage{epsfig}
\usepackage{graphicx}
\usepackage{amsmath}
\usepackage{amssymb}
\usepackage{breqn}
\usepackage{multirow}

\ifCLASSOPTIONcompsoc
  \usepackage[nocompress]{cite}
\else
  \usepackage{cite}
\fi
\ifCLASSINFOpdf
\else
\fi
\begin{document}
%
\title{Inter-Homines: Distance-Based Risk Estimation for Human Safety}
%
%
%
%

\author{Matteo~Fabbri, Fabio~Lanzi, Riccardo~Gasparini, Simone~Calderara, Lorenzo~Baraldi, Rita~Cucchiara
\IEEEcompsocitemizethanks{\IEEEcompsocthanksitem The authors are with the Department of Engineering ``Enzo Ferrari'', University of Modena and Reggio Emilia, Italy. \protect\\
E-mail: name.surname@unimore.it.}
}

\IEEEtitleabstractindextext{%
\begin{abstract}
  In this document, we report our proposal for modeling the risk of possible contagiousity in a given area monitored by RGB cameras where people freely move and interact. Our system, called Inter-Homines, evaluates in real-time the contagion risk in a monitored area by analyzing video streams: it is able to locate people in 3D space, calculate interpersonal distances and predict risk levels by building dynamic maps of the monitored area. Inter-Homines works both indoor and outdoor, in public and private crowded areas. The software is applicable to already installed cameras or low-cost cameras on industrial PCs, equipped with an additional embedded edge-AI system for temporary measurements. From the AI-side, we exploit a robust pipeline for real-time people detection and localization in the ground plane by homographic transformation based on state-of-the-art computer vision algorithms; it is a combination of a people detector and a pose estimator. From the risk modeling side, we propose a parametric model for a spatio-temporal dynamic risk estimation, that, validated by epidemiologists, could be useful for safety monitoring the acceptance of social distancing prevention measures by predicting the risk level of the scene.
\end{abstract}

}

\maketitle

\IEEEdisplaynontitleabstractindextext

%
\IEEEpeerreviewmaketitle

\begin{figure*}[t!]
\begin{center}
\includegraphics[width=0.95\linewidth]{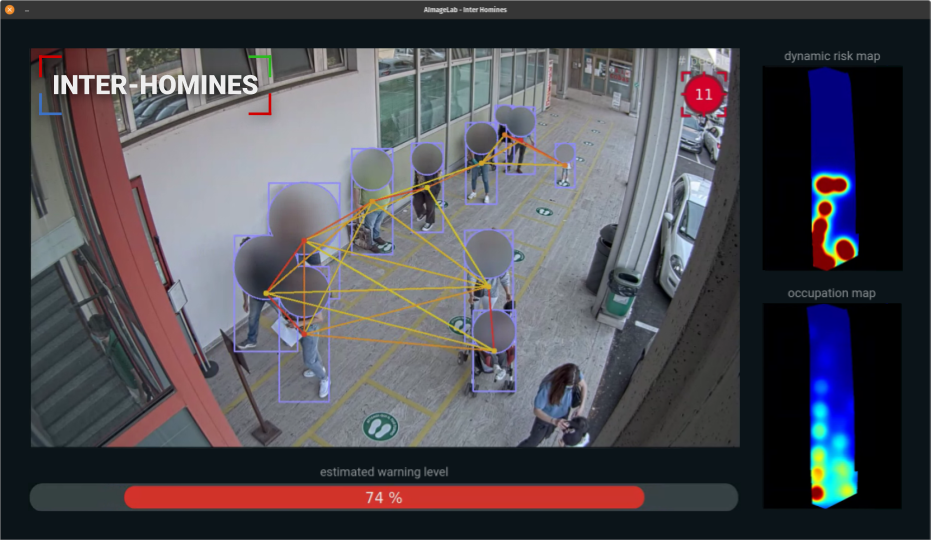}
\end{center}
    \vspace{-10px}
    \caption{GUI of our system. In the main frame, anonymized bounding boxes are superimposed to the image. Colored links encodes people reciprocal distance. On the right, two maps shows the bird-eye view of the area. The estimated risk level of the scene resides at the bottom of the interface.}
\label{fig:gui}
\end{figure*}

\IEEEraisesectionheading{\section{Introduction}\label{sec:introduction}}
\IEEEPARstart{T}{he} COVID-19 emergency has changed the way we live interpersonal social relationships, at work, in public and private spaces, in places of education culture and leisure. The risk of contagion seems full-blown; until now, there are no conclusive studies which correlate environmental and endogenous factors with the greatest spread of the virus: instead, everything seems to correlate the contagion to proximity or to the contact between infected people and people susceptible to infection \cite{asadi2020coronavirus}. The spread of the infection seems to follow the epidemiological models that derive from the SIR models \cite{chen2020time}.

The phases that all the world is going to undertake after the lock-down will be characterized by living with the risk of contagion: the prerogative will be to take conscious and possibly interactive measures to minimise the possibility of contagion, while seeking a necessary resumption of social and working life. 

Certainly, the IT technologies and in particular Artificial Intelligence can be valuable tools to monitor and predict risk levels in potentially crowded places. In fact, we propose an innovative and effective technological contribution based on Computer Vision and Deep Learning, in order to dynamic monitoring the acceptance of social distancing prevention measures through real-time calculation of the risk level, with particular reference to workplaces, public places and social areas. For statistical purposes, people behavior dynamics are stored in a database in a completely privacy compliant manner. The data can be used to identify the most critical areas and hours of the day in terms of number of people and risk level, in order to better address distance-related interpersonal prevention measures. 

The system is called Inter-Homines (from the ``Homo inter homines sum, capite aperto ambulo'' - ``I am a human among humans and I can walk with my face uncovered'') because people should be free to move and interact with uncovered faces while being safe at the same time.

The system has a twofold goal. The first is to provide a reliable tool, in accordance with European privacy and usage guidelines of the AI, to calculate in real time the actual compliance with the prevention measures for "spacing", also interactively reporting any risky situations. In particular, the implemented system can generate real-time alarms when people form crowds. The second goal is to provide an innovative model for the dynamic calculation of the risk of the monitored site that can be used as a tool for prevention, control, monitoring, and planning, support to the population and workers in order to implement conscious attendance, linked to effective compliance with the measures in force.
 
Detecting people, their position in the space, their mutual distance is a typical application of Computer Vision. Many tools are available, using state-of-the-art deep learning architecture and geometry-based 3D reconstruction. Results are promising although still far to be applied everywhere by everyone. In this project, we can take advantage of a long term experience in computer vision for surveillance and people behavior understanding \cite{fabbri2018learning, fabbri2020compressed}, providing a novel detection pipeline running in real-time. It exploits standard fast camera calibrations, a people detector and a pose estimation methods.

Inter-Homines defines a model, validated by epidemiologists and parameterizable according to current regulations, which allows, in real-time, to associate each monitored area with: a) a space-time risk index, b) a dynamic safety level of the area, c) a dynamic map of interpersonal distances and d) a real time visualization of detected persons and distances. See Fig.~\ref{fig:gui} for the system output overview.

\section{Related Work}
One of the most popular two-stage deep object detectors is R-CNN \cite{girshick2014rich} which predicts object location from a set of region proposals \cite{xiang2017subcategory}, crops them and classifies each using a second deep neural network. Fast R-CNN \cite{girshick2015fast}, instead, directly crops image features to save computation. However, both approaches rely on slow low-level region proposal techniques.

On the other hand, one-stage methods such as Faster R-CNN \cite{ren2015faster} generates region candidates within the detection network. It samples bounding boxes with fixed shape (anchors) around the image grid and classifies them into foreground or background. Each proposal is then further classified into object classes. Several improvements to one-stage detectors include anchor shape priors such as in YOLO \cite{redmon2017yolo9000,redmon2018yolov3}, SSD's different feature resolution \cite{liu2016ssd}, and loss re-weighting among different samples \cite{lin2017focal}.

Our approach leverages CenterNet \cite{zhou2019objects}, which is closely related to anchor-based one-stage detectors. However, CenterNet does not requires manual thresholds for foreground and background classification and does not require Non-Maximum Suppression (NMS) \cite{bodla2017soft} post processing as it simply extracts local peaks in the keypoint heatmap \cite{cao2017realtime, fabbri2018learning}. Moreover, CenterNet utilizes an output stride of 4 which is 2 times larger than in traditional object detectors \cite{he2016deep, he2017mask}, making it more accurate.

Other methods utilize the same robust keypoint estimation network as CenterNet: CornerNet \cite{law2018cornernet} and ExtremeNet \cite{zhou2019bottom}. CornerNet detects the bounding box corners as keypoints while ExtremeNet predicts the left, top, right and bottom extremes of the objects. However, those methods require a combinatorial grouping stage as post processing, which considerably slows down the whole pipeline. CenterNet, instead, simply extracts a single center point per object without the need for grouping or post-processing.

People detection can be also achieved by pose estimation. The trend of pose estimation \cite{cao2017realtime, newell2016stacked,insafutdinov2016deepercut} is very promising, but often is too computational severe to be implemented for real time edge applications with an unknown number of people. Thus, in this work, we adopt a simplified pose estimation algorithm, that it is used together with the people detector to make the localization more robust to occlusions.

Many 3D object detection methods have been proposed in literature. Among them, 3D R-CNN \cite{kundu20183d} adds a further head to Faster R-CNN \cite{ren2015faster} which is followed by a 3D projection. Also Deep Manta \cite{chabot2017deep} exploits a coarse-to-fine Faster R-CNN \cite{ren2015faster} trained on multiple tasks. Finally, Deep3Dbox \cite{mousavian20173d} utilizes a slow R-CNN \cite{girshick2014rich} by first predicting 2D bounding boxes and then feeding each detection into a 3D estimation network. However, those methods require huge computational power and does not leverage constraints such as fixed camera and flat ground plane.

\section{$\mathcal{R}_0$ and the SIR model}
After the outbreak of the COVID‑19 pandemic, all the world learned the importance of the basic reproduction number, $\mathcal{R}_0$, as the statistical index indicating the degree of spread of the infection. In commonly used infection models, when $\mathcal{R}_0>1$ (in Italy has reached 4.3 during the spring of 2020) the infection will be able to start spreading in a population, but not if $\mathcal{R}_0<1$. Generally, the larger the value of $\mathcal{R}_0$, the harder it is to control the epidemic.

$\mathcal{R}_0$ is defined as the expected number of secondary cases produced by an infection in a completely susceptible population:
\begin{equation}
    \mathcal{R}_0 = \alpha \cdot c \cdot d
    \label{eq:rr}
\end{equation}
where $\alpha$ is the transmissibility, $c$ is the average rate of contact between susceptible and infected individuals, and $d$ is the duration of infectiousness.

To understand if this quantity defines the epidemic threshold of a particular infection, we need to formulate a Susceptible-Infected-Removed (SIR) epidemic model \cite{kermack1927contribution}. This model deploys several assumptions: 1) closed population, 2) constant rates, 3) no births and deaths and 4) well mixed population.
\begin{figure*}[t!]
\begin{center}
\includegraphics[width=0.95\linewidth]{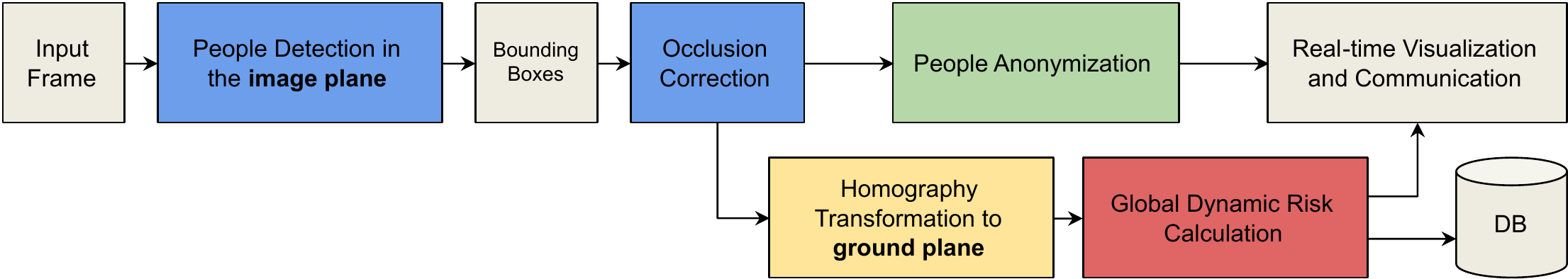}
\end{center}
    \vspace{-10px}
    \caption{Schematization of the Inter-Homines pipeline: the input frame is processed to produce bounding box detections. Each detection is then refined by the Occlusion Correction module that copes with truncated bounding boxes. The image plane detection coordinates are then transformed to ground plane coordinates using an Homography Transformation. Those coordinates are then used to calculate the global risk. People coordinates and risk level are then stored into a database. Finally, the system outputs the anonymized frame along with the risk level and the risk maps.}
\label{fig:main}
\end{figure*}

Given a population of N individuals, let's consider $S$ the number of susceptible people, $I$ the infected, and $R$ the removed. Removed people are those that cannot be infected, as they might have developed antibodies. Now let's define $s=\frac{S}{N}$, $i=\frac{I}{N}$, $r=\frac{R}{N}$ as the fraction in each set. The SIR model is defied as:
\begin{equation}
    \frac{ds}{dt} = -\beta si ,\quad\quad \frac{di}{dt} = \beta si - vi ,\quad\quad \frac{dr}{dt} = vi
    \label{eq:R}
\end{equation}




An epidemic occurs if the number of infected increases: $\frac{di}{dt}>0$. By considering that everyone is susceptible, we can substitute $s=1$ obtaining the following inequality:
\begin{equation}
    \alpha c d = \mathcal{R}_0 > 0
    \label{eq:R}
\end{equation}

$\mathcal{R}_0$ is essentially the entire theoretical basis of public health interventions for infectious diseases and it is simply the product of the transmissibility, the mean contact rate, and the duration of infection. In order to reduce transmissibility $\alpha$ we can develop vaccines, get people to use barrier contraceptives or use anti-retrovirals. To decrease mean contact $c$, the world decided to use isolation/quarantine, and  health education programs. Finally, to reduce the duration of infection $d$ therapeutics, antibiotic treatment of bacterial infections that boost innate immune response can be exploited.

$\mathcal{R}_0$ is in generally computed as a posterior measure, but cannot be dynamically predicted in a robust way since the factor influencing $\mathcal{R}_0$ are not a priori easily measurable. In this work we cannot do anything a part from monitoring the acceptance of health education programs. In the past months, many countries decided the mandatory measures of security that concern the use of DPI and the social distance guidelines. Thus, in order to make $c$ as small as possible, we should keep all the people at a distance larger that a threshold distance of a possible infection.

A viable way is to force people to stay in queues, maybe with some marker placed on the floor and with the constant attention of a human guard that controls the compliance of the social distancing norms. This is not always possible, especially in big malls and wide areas. Moreover, the human monitoring is not always optimal as the guard is subject to tiredness and lost of focus.

This is the reason why computer based systems joined with risk models can substitute human controllers and help to perform real-time monitoring of areas by assessing a level of possible risk, and, if necessary, giving a real time feedback to improve the safety and decrease the risk. In the following section, we propose a very simple model, that, using some thresholds validated by epidemiologists, models the dynamic risk in a given area.

\section{Risk Model}
The SIR model formulation, as described in the previous section, has validity when considering a population. Now, let's consider a much more restricted zone. This could be an indoor area, like a waiting room of a public office, an entrance in a cinema or a shop. More precisely, let's consider a scene with $N$ people $k_0, \dots, k_{N-1}$ at a given time $t$. Given two subjects $k_i$ and $k_j$ with distance $d_{i,j}$, we define their reciprocal risk as follows:
\begin{equation}
    rr_{i,j}^{(t)} = \eta \, e^{-\beta \max(0, \, d_{i,j} - \tau)}
    \label{eq:rr}
\end{equation}
where $\eta$, $\beta$ and $\tau$ are parameters that respectively control height, slope and the full width at maximum of the function. In this specific application, $\eta$ is a mitigator used to decrease the risk when some criteria are met, e.g., when at least one of the two people is wearing a facial mask. $\beta$, instead, controls how the risk decreases when the distance is greater than $\tau$ and can model environmental characteristics such as air temperature and the presence of air conditioning. Lastly, $\tau$, controls the transmissibility of the disease via respiratory droplets and define the minimal distance allowed between two people. It should follow World Health Organization and national guidelines but can be further increased to better preserve the safety of people in critical places such as COVID-19 hospital units. We then define the individual risk at time $t$ as:
\begin{equation}
    R_i^{(t)} = \max_{j=0 \dots N-1, j \neq i}\{ rr_{i,j}^{(t)} \}
    \label{eq:R}
\end{equation}
The global risk at $t$ of the scene is then computed as follows:
\begin{equation}
    G^{(t)} = \min \Big(1, \; \frac{1}{C} \sum_{i=0}^{N-1} R_i^{(t)}\Big)
    \label{eq:G}
\end{equation}
where $C$ is the maximum capacity of the scene. This capacity can be either given by the user or calculated using simple covering algorithms. Finally. the dynamic global risk is computed as:
\begin{equation}
    D^{(t)} = \frac{1}{W} \sum_{w=0}^{W-1} G^{(t-w)}
    \label{eq:D}
\end{equation}
where $W$ is the size of the temporal window. At a given time $t$, $D^{(t)} \in [0, 1]$ is the global risk of the scene and it is used to trigger alarms when it reaches a given threshold.

\section{Inter-Homines Technical Core}
Here we give an overview of the pipeline we used to process videos in real time. The aim of our Inter-Homines system is to detect people, compute their distance and provide a dynamic risk level of the area, as well as producing a human readable visualization with anonymized people. For GDPR constraints, no visual data is recorded but, instead, only people coordinates are extracted and stored. Data is acquired with a variable rate, up to one time per second for each camera. See Fig.~\ref{fig:main} for a schematization of the pipeline.

The following subsections summarize the key elements of our system. Section~\ref{sec:people} describes the people detection stage and elaborates on its challenges. Section~\ref{sec:pose} illustrates our proposed keypoint localization solution which addresses the occulsion problem peculiar of surveillance scenarios and also provide the head position for anonymization purposes. Next, in Section~\ref{sec:homography}, we describe how we convert points from image plane to ground plane and, finally, Section~\ref{sec:output} illustrates the system outputs.

\subsection{People Detection}
\label{sec:people}
As we are interested in the best speed-accuracy trade-of, we choose CenterNet \cite{zhou2019objects} as a people detector. In particular, we rely on the DLA backbone \cite{yu2018deep} which yields 51.3\% AP for the people class on MS COCO \cite{lin2014microsoft}, running at 52 FPS on a Titian XP.

Let $I \in R^{W \times H \times 3}$ be the input image having width $W$ and height $H$. CenterNet outputs a keypoint heatmap $\hat Y \in [0,1]^{\frac{W}{R} \times \frac{H}{R} \times C}$, where $R=4$ is the output stride and $C$ is the number of keypoint types. Keypoint types include $C=80$ object categories but in this work we only consider the ``people'' class.
Detected keypoints corresponds to a prediction $\hat Y_{x,y,c}, = 1 $ and $0$ otherwise. To recover the discretization error generated by the output stride, CenterNet further predicts a local offset $\hat O \in \mathcal{R}^{\frac{W}{R} \times \frac{H}{R} \times 2}$ for each center point.

Let $(x_1^{(k)}, y_1^{(k)}, x_2^{(k)}, y_2^{(k)})$ be the bounding box of object $k$ of the ``people'' class and $p_k = (\frac{x_1^{(k)} + x_2^{(k)}}{2}, \frac{y_1^{(k)} + y_2^{(k)}}{2})$ it's center point.
CenterNet predicts all center points for each object $k$ and further regresses to the object size $s_k = (x_2^{(k)} - x_1^{(k)}, y_2^{(k)} - y_1^{(k)})$.

At running time, we first extract the peaks in the heatmap for the ``people'' category. We consider all the responses whose value is greater or equal to its 8-connected neighbors. Let  $\hat{\mathcal{P}} = \{(\hat x_i, \hat y_i)\}_{i = 1}^{n}$ be the set of $n$ detected center points where keypoint values $\hat Y_{x_iy_ic}$ are utilized as a measure of its detection confidence. Bounding boxes are produced at location:
\begin{equation} \label{eq1}
\begin{split}
(&\hat x_i + \delta \hat x_i - \hat w_i / 2,\ \ \hat y_i + \delta \hat y_i - \hat h_i / 2, \\
&\hat x_i + \delta \hat x_i + \hat w_i / 2,\ \ \hat y_i + \delta \hat y_i + \hat h_i / 2),
\end{split}
\end{equation}
where $(\delta \hat x_i, \delta \hat y_i) = \hat O_{\hat x_i,\hat y_i}$ is the predicted offset and $(\hat w_i,\hat h_i) = \hat S_{\hat x_i,\hat y_i}$ is the predicted size.
Since the prediction are directly produced from the keypoint estimation, there is no need for IoU-based NMS or other post-processing techniques. This makes CenterNet faster w.r.t. other detectors, making it suitable for real time applications. 

CenterNet is capable of producing a precise localization of every person in the image, however, it does not take into account occlusions that usually happen in real world scenarios. As shown in Fig.~\ref{fig:talpa} (pink bounding boxes), if a person is occluded by an object or by other people, CenterNet predicts a tight bounding box that only contains the visible part of the person, ignoring his full shape. This usually happens with the bottom part of the body, as the camera is commonly placed several meters above the ground. Since we are ultimately interested in recovering the ground plane coordinate of each person through homograpy, we need to know the exact position (in image plane) of the feet of each detected person. This task cannot be accomplished by solely relying on CenterNet.

\subsection{Feet and Head Localization}
\label{sec:pose}
To overcome the aforementioned limitations without introducing complexity to the overall system, we propose to utilize a small network to predict the feet position given a bounding box containing a person, even if the feet are not visible.

To this aim we rely on a simple but effective CNN that, given an image $M$ tightly containing a person, it regresses to the midpoint $P_f = (x_f, y_f)$ of the segment having the two feet as endpoints. This ensures that we know the exact position in image plane where every person touches the ground. Since we are also interested in anonymizing the face of each detected person, we further predict the location of the head $P_h = (x_h, y_h)$.

We replaced the last 1000 class classification layer of Resnet50 \cite{he2016deep} with two heads composed by an adaptive average pooling layer and a fully connected layer with output size equal to 2. The adaptive average pooling takes care of the difference in size that each bounding box fed to the network can have. Training has been carried out for 10 epochs using an MSE loss with Adam optimizer, batch size of 64 and learning rate 0.001.

\begin{figure}[t!]
\begin{center}
\includegraphics[width=0.95\linewidth]{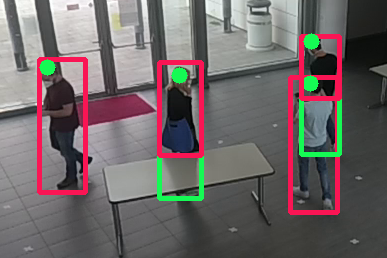}
\end{center}
    \vspace{-10px}
    \caption{Examples of CenterNet bounding boxes (pink), refined bounding boxes and head localization (green). }
\label{fig:talpa}
\end{figure}

We used JTA \cite{fabbri2018learning} as the training dataset since it is the only surveillance dataset available in literature that provide pose estimation annotations with occlusion information. Thanks to this, we were able to simulate occlusion situations by simply picking, during training, the pedestrians with the bottom keypoints occluded, like ankles, knees, and hips. During training, we also randomly shortened some of the bounding boxes in order to simulate CenterNet behaviours. This step ensures a more precise localization of the feet while also coping with truncated bounding boxes. As shown in Fig.~\ref{fig:talpa} (green bounding boxes), our network can effectively obtain an accurate position of each head and it is used to extend the bounding box to its regular shape.

\begin{figure}[t!]
\begin{center}
\includegraphics[width=0.49\linewidth]{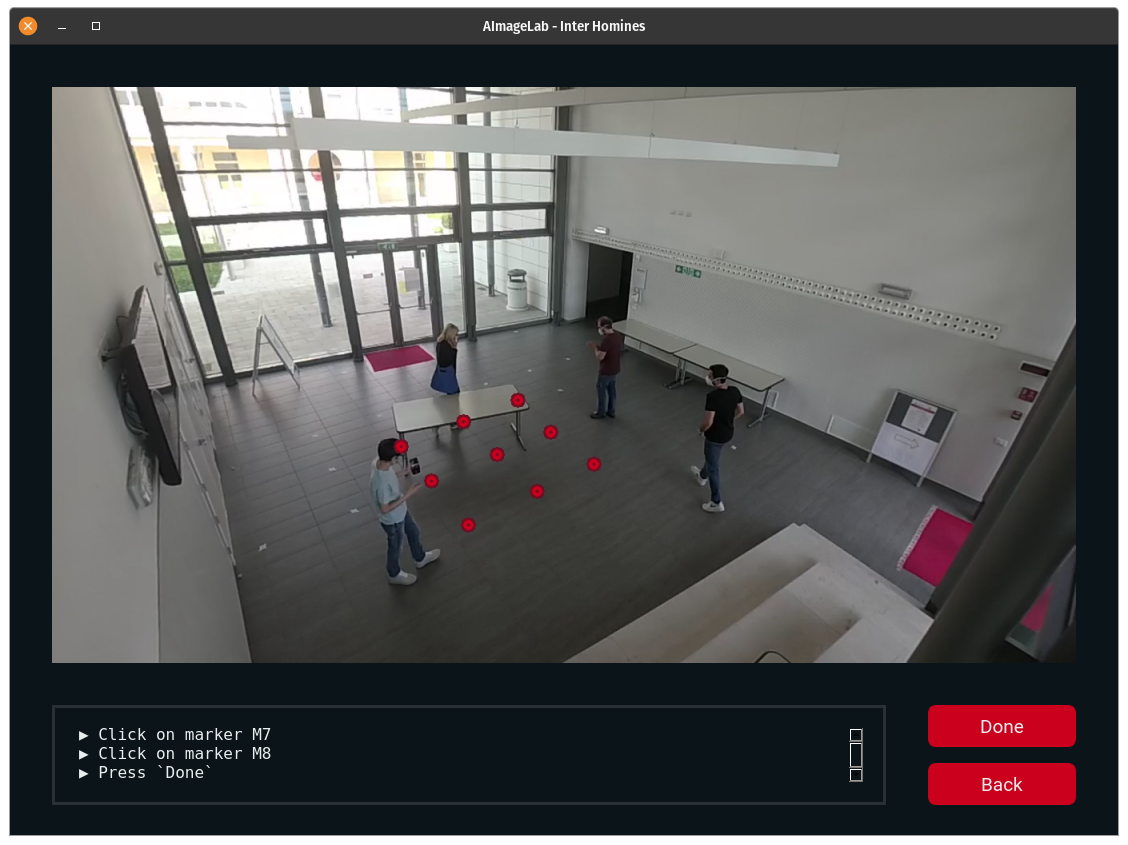}
\includegraphics[width=0.49\linewidth]{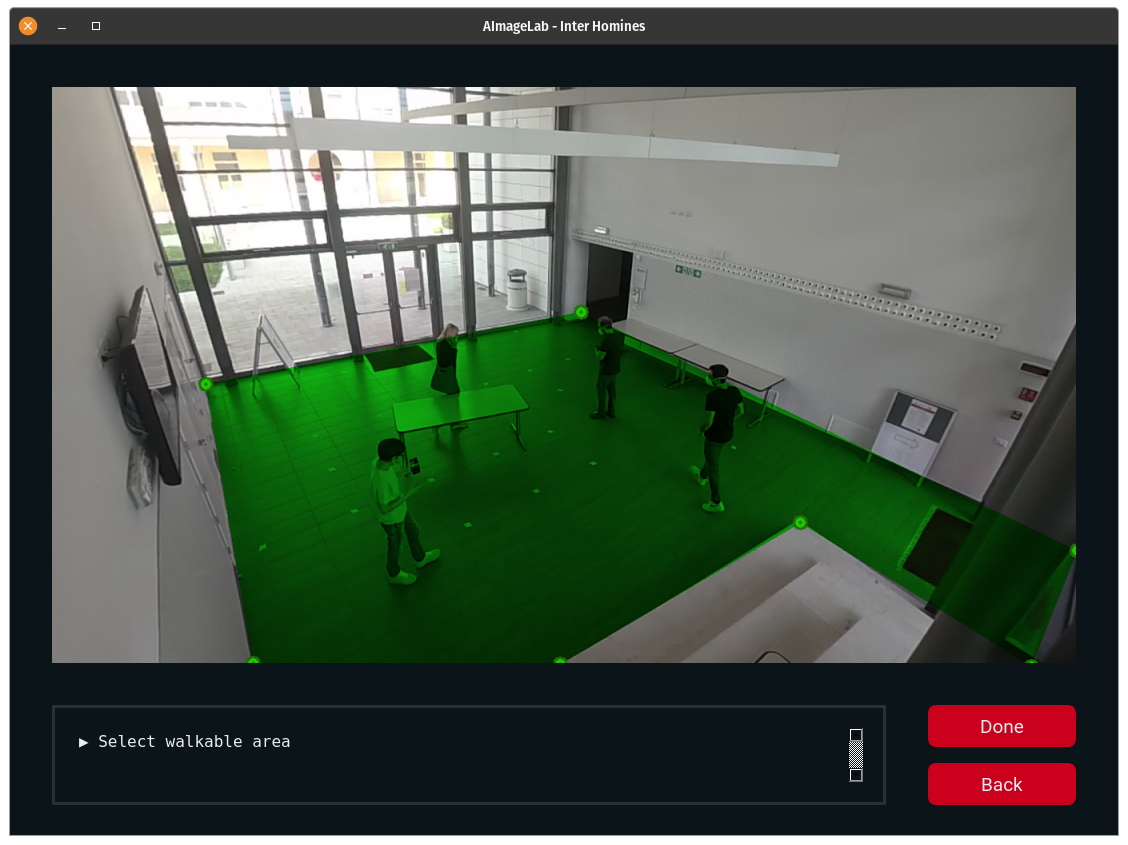}
\end{center}
    \vspace{-10px}
    \caption{GUI used during system calibration for homography matrix calculation (left) and for walking area selection (right).}
\label{fig:i2}
\end{figure}

\begin{figure*}[t]
\begin{center}
\includegraphics[width=0.33\textwidth]{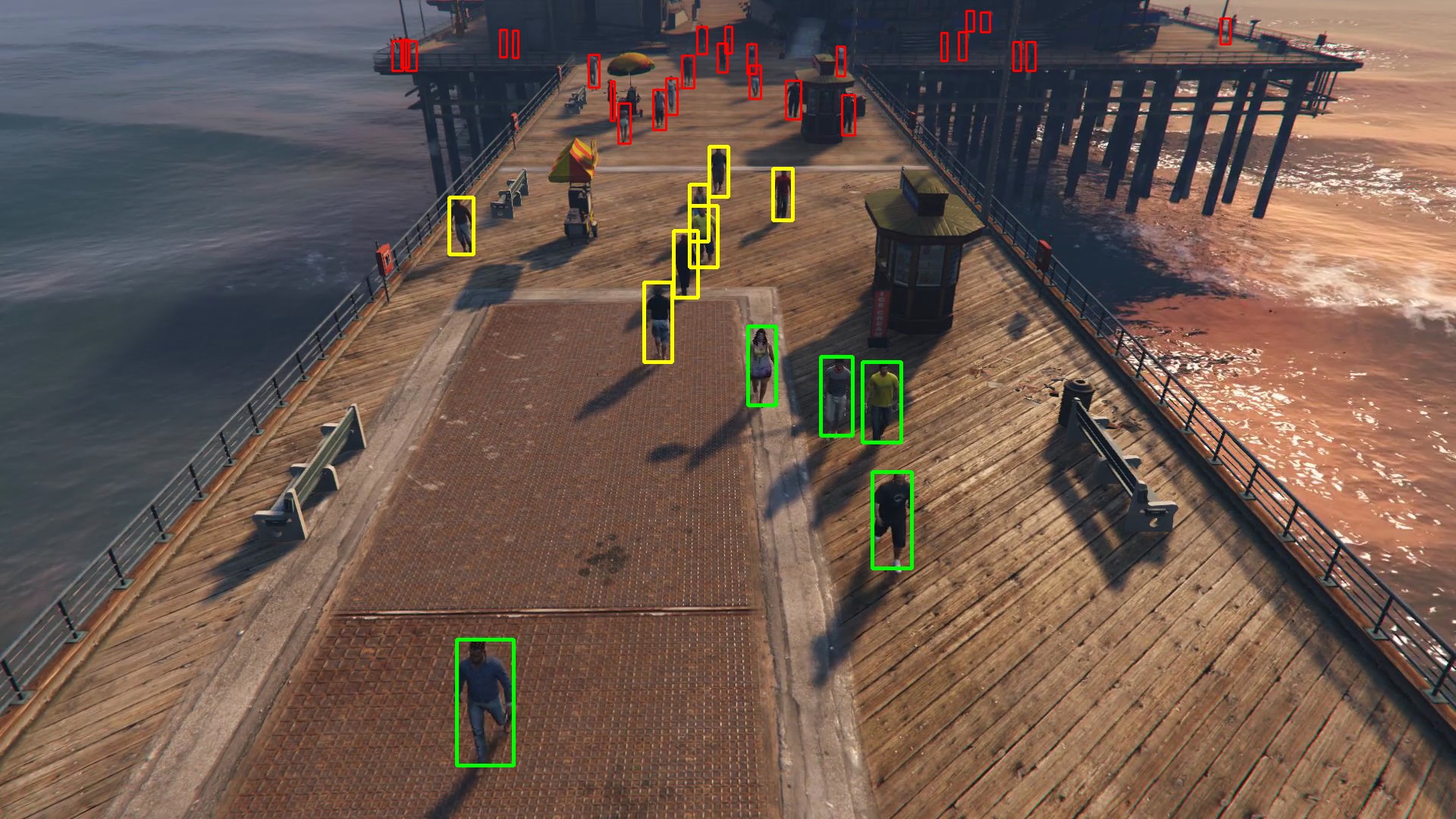}
\includegraphics[width=0.33\textwidth]{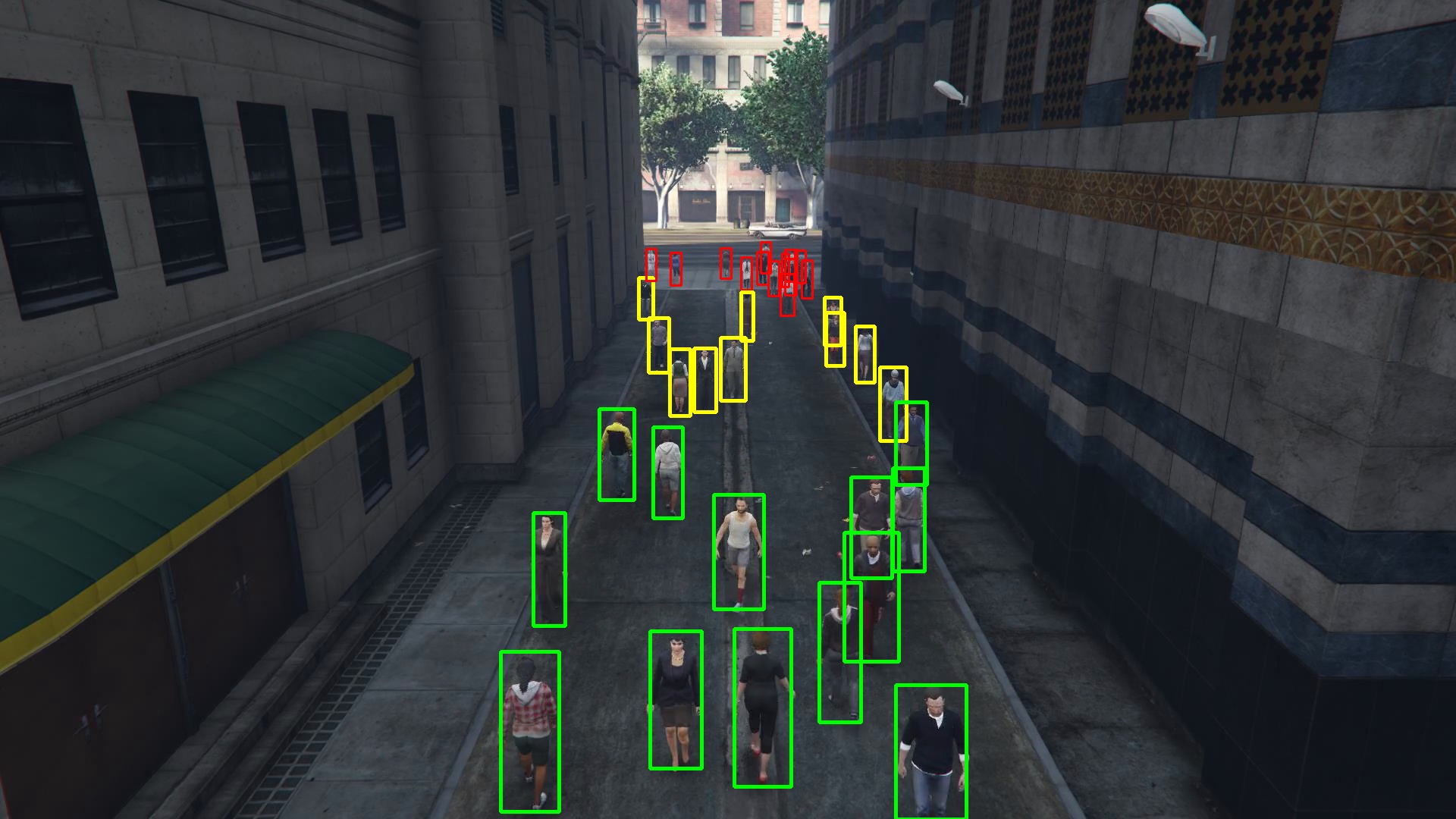}
\includegraphics[width=0.33\textwidth]{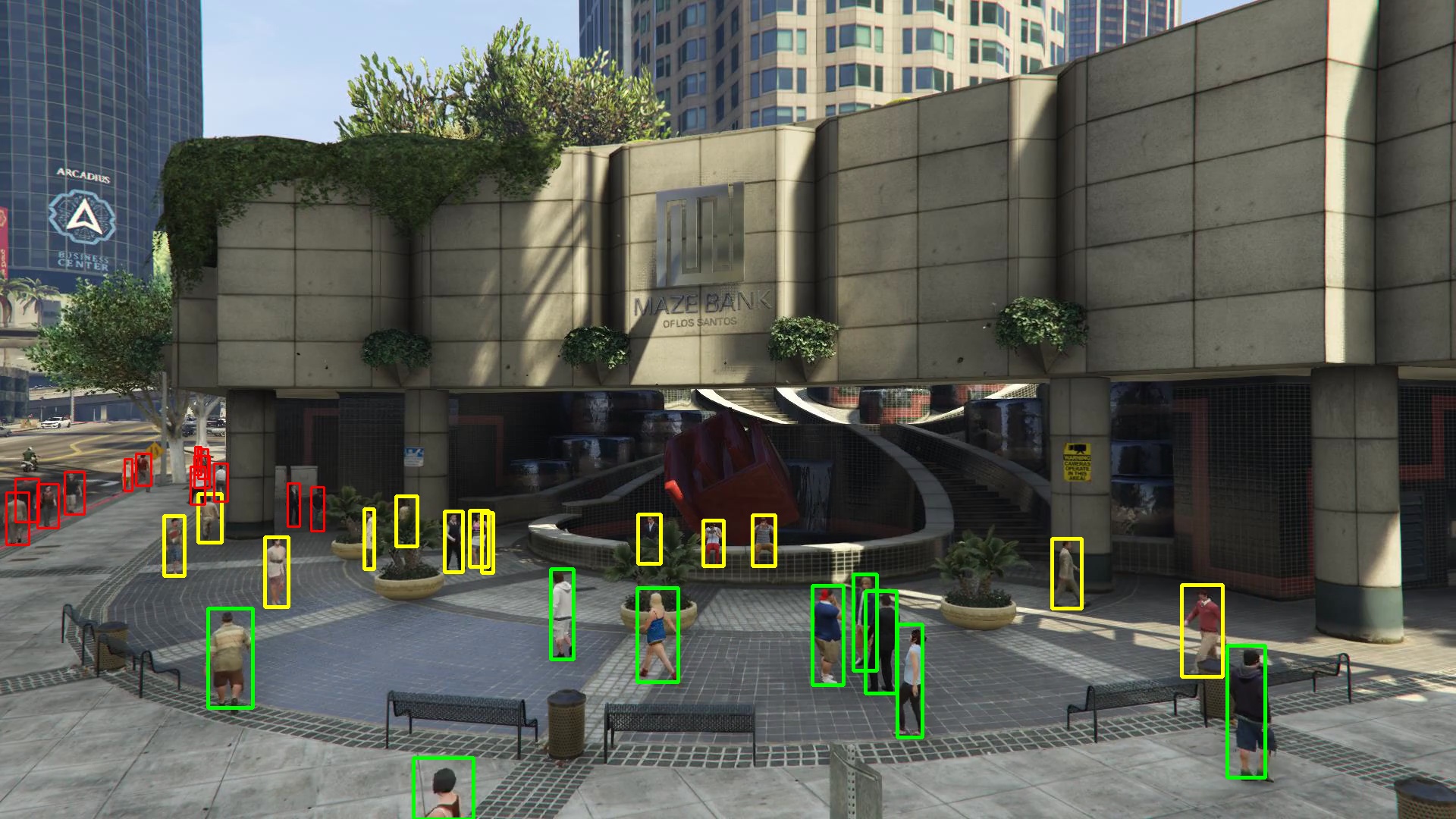}
\end{center}
\caption{Examples from the JTA dataset exhibiting its variety in viewpoints, number of people and scenarios. Ground truth bounding boxes are superimposed to the original images. Green color is used for people having a distance from the camera between 0 and 20 meters, yellow for people between 20 and 40 meters and red for people between 40 and 100 meters.}
\label{fig:JTA}
\end{figure*}

\subsection{From Image Plane to Ground Plane}
The camera projection matrix $P$ is a $3\times 4$ matrix which describes the mapping of a pinhole camera \cite{hartley2003multiple} from 3D points in the world to 2D points in an image. Let $X$  be a representation of a 3D point in homogeneous coordinates (a 4-dimensional vector), and let $y$ be a representation of the image of this point in the pinhole camera (a 3-dimensional vector), we have $y=PX$. The camera projection matrix can be decomposed as:
\label{sec:homography}
\begin{equation}
\begin{bmatrix}
u\\ 
v\\ 
1
\end{bmatrix}
=
\begin{bmatrix}
f_x & 0 & c_x & 0\\ 
0 & f_y & c_y & 0\\ 
0 & 0 & 1 & 0
\end{bmatrix}
\begin{bmatrix}
r_{1,1} & r_{1,2} & r_{1,3} & t_X\\ 
r_{2,1} & r_{2,2} & r_{2,3} & t_Y\\ 
r_{3,1} & r_{3,2} & r_{3,3} & t_Z\\
0 & 0 & 0 & 1
\end{bmatrix}
\begin{bmatrix}
X\\ 
Y\\ 
Z\\
1
\end{bmatrix}
\label{HHHH}
\end{equation}
where the intrinsic parameters $f_x$, $f_y$ and $c_x$, $c_y$ are the camera focal length and principal points respectively while $r_{i,j}$ and $t_i$ are the extrinsic parameters which define the rotation and the translation used to describe the rigid motion of an object in front of a still camera. Finally, $u$ and $v$ are the coordinates of the projected point in pixels while $X$, $Y$ and $Z$ are the coordinates of a 3D point in the world coordinate space. By considering the simpler case of a projection of planar points, where each 3D point lies on the same plane (e.g. the ground), we can simplify the formulation considering $Z=0$. For planar surfaces, 3D to 2D perspective projection reduces to a 2D to 2D transformation:
\label{sec:homography}
\begin{equation}
\begin{bmatrix}
u\\ 
v\\ 
1
\end{bmatrix}
=
\begin{bmatrix}
f_x & 0 & c_x\\ 
0 & f_y & c_y\\ 
0 & 0 & 1
\end{bmatrix}
\begin{bmatrix}
r_{1,1} & r_{1,2} & t_1\\ 
r_{2,1} & r_{2,2} & t_2\\ 
r_{3,1} & r_{3,2} & t_3\\
\end{bmatrix}
\begin{bmatrix}
X\\ 
Y\\ 
1
\end{bmatrix}
\label{HH}
\end{equation}
and by doing the products we finally obtain the planar homography matrix $H$. The planar homography relates the transformation between two planes (e.g. the image plane and the ground plane):
\begin{equation}
\begin{bmatrix}
u\\ 
v\\ 
1
\end{bmatrix}
=
\begin{bmatrix}
h_{1,1} & h_{1,2} & h_{1,3}\\ 
h_{2,1} & h_{2,2} & h_{2,3}\\ 
h_{3,1} & h_{3,2} & h_{3,3}
\end{bmatrix}
\begin{bmatrix}
X\\ 
Y\\ 
1
\end{bmatrix}
=
H
\begin{bmatrix}
X\\ 
Y\\ 
1
\end{bmatrix}
\label{eq:H}
\end{equation}
Since $H$ maps from ground plane to image plane, but we are interested in the opposite transformation (from image plane to ground plane), we now need to calculate the inverse homography matrix $H^{-1}$. An homography matrix $H$ is always invertible, and its inverse is still an homography transformation:
\begin{equation}
\begin{bmatrix}
X\\ 
Y\\ 
1
\end{bmatrix}
=
H^{-1}
\begin{bmatrix}
u\\ 
v\\ 
1
\end{bmatrix}
\label{eq:H-1}
\end{equation}
A practical way to calculate the homography matrix $H$ of Eq.~\ref{eq:H} is to find a set of at least four points pairs of target and source planes and to minimize the back-projection error:
\begin{dmath}
\sum_{i=0}^N \Bigg[\bigg(u_i - \frac{h_{1,1}X_i+h_{1,2}Y_i+h_{1,3}}{h_{3,1}X_i+h_{3,2}Y_i+h_{3,3}}\bigg)^2 + 
\bigg(v_i - \frac{h_{2,1}X_i+h_{2,2}Y_i+h_{2,3}}{h_{3,1}X_i+h_{3,2}Y_i+h_{3,3}}\bigg)^2\Bigg]
\label{back}
\end{dmath}
\begin{table*}[t!]
\small
\centering
\caption{3D detection results on JTA Dataset. In PR (precision), RE (recall) and F1, @$t$ indicates that a predicted person is considered ``true positive'' if the distance from the corresponding ground truth location is less than $t$. The max range indicates the maximum distance considered in the calculation.}
\begin{tabular}{clccccccccccccccc}
\multicolumn{2}{c}{}      &  & PR    & RE    & F1   &  & PR    & RE    & F1   &  & PR    & RE    & F1   
\\ \cline{4-6} \cline{8-10} \cline{12-14} \\[-1em]
max range & \multicolumn{1}{c}{} 
 &  & \multicolumn{3}{c}{@0.5 m}                                            
 &  & \multicolumn{3}{c}{@1.0 m}       
 &  & \multicolumn{3}{c}{@1.5 m} \\
\hline
\\[-0.7em]
\multirow{2}{*}{10m} & Inter-Homines w/o Occ. Corr.
                       &  & 83.38  &  78.29  &  80.06
                       &  & 90.44  &  85.28  &  87.07
                       &  & 92.56  &  87.63  &  89.31
                       \\
& Inter-Homines Full Pipeline
                       &  & 88.01  &  84.74  &  85.55
                       &  & 92.95  &  89.2  &  90.22
                       &  & 94.27  &  90.56  &  91.59
                       \\[0.2em]
\hline
\\[-0.7em]
\multirow{2}{*}{20m} & Inter-Homines w/o Occ. Corr.
                       &  & 69.11 & 59.75 & 63.41 
                       &  & 85.33 & 73.86 & 78.36 
                       &  & 91.83 & 79.77 & 84.48 
                       \\
& Inter-Homines Full Pipeline
                       &  & 74.96 & 66.98 & 70.00 
                       &  & 88.86 & 79.20 & 82.87 
                       &  & 93.64 & 83.69 & 87.49 
                       \\[0.2em]
\hline
\\[-0.7em]
\multirow{2}{*}{30m} & Inter-Homines w/o Occ. Corr.
                       &  & 59.47  &  46.87  &  51.57
                       &  & 77.39  &  60.76  &  66.96
                       &  & 85.81  &  67.27  &  74.14
                       \\
& Inter-Homines Full Pipeline
                       &  & 65.3  &  53.07  &  57.52
                       &  & 81.26  &  65.4  &  71.18
                       &  & 88.34  &  70.96  &  77.31
                       \\[0.2em]
\hline
\\[-0.7em]
\multirow{2}{*}{100m} & Inter-Homines w/o Occ. Corr.
                       &  & 53.76  &  31.65  &  38.07
                       &  & 71.34  &  41.74  &  50.41
                       &  & 80.3  &  46.88  &  56.7
                       \\
& Inter-Homines Full Pipeline
                       &  & 60.91  &  36.21  &  43.37
                       &  & 77.12  &  45.21  &  54.55
                       &  & 84.75  &  49.41  &  59.79
                       \\[0.2em]
\hline
\end{tabular}
\label{tab:cp_exp_jta}
\end{table*}
However, if not all of the point pairs fit the rigid perspective transformation, this initial estimate will be poor. To solve this problem we employ the RANSAC iterative method, trying many different random subsets of the corresponding point pairs (of four pairs each). We then estimate the homography matrix applying a simple least-square algorithm using this subset, and then compute the quality of the computed homography, which is the number of inliers. The best subset is then used to produce the initial estimate of the homography matrix. The computed homography matrix is refined further (using only the inliers) with the Levenberg-Marquardt method to further reduce the re-projection error. The homography matrix is determined up to a scale. Thus, it is normalized so that $h_{3,3}=1$.

This method of using an homography transformation to obtain 3D coordinates is the most appropriate when we want to monitor an approximately flat area (such as a town square) using a fixed camera and there is the possibility of making appropriate measurements in the monitored space. 

To easily obtain the points pairs of image and ground planes needed to find the homography matrix $H$, we designed a simple procedure that we call ``system calibration''. This procedure consists in placing nine markers at the center of the monitored area, fully visible from the camera. The markers are placed in a grid pattern as in Fig.~\ref{fig:i2}. By means of a simple graphic interface, the user can take a snapshot of the camera and click with the cursor the centers of the nine markers in order to acquire the pixel coordinates. In practice, we utilize a special carpet with the nine markers printed on it. The use of the carpet automates the real world measurements as we already know the distance between markers in the carpet, making the system calibration fast, less prone to errors and feasible by everyone. Once the nine pairs of points have been identified and the homography matrix calculated, the carpet can be safely removed and the system will continue to work properly as long as the camera maintains its position.

During the system calibration an optional procedure of selecting the ``walking area'' can be carried out. Again, a simple graphic interface let the user draw a polygon on the snapshot taken from the camera, as shown in Fig.~\ref{fig:i2}. The pixel vertices are then converted to ground coordinates that are used to exclude detections whose 3D position is outside the walking area. This is useful, for example, to ignore mirrors or windows that can reflect people causing unwanted detections. 

Given a bounding box of a person, we can now extract its central point $(u,v)$ of the lower side of the box (i.e. the image coordinate where the person touches the ground with his feet), and utilize Eq~\ref{eq:H-1} to obtain the corresponding $(X,Y)$ point in ground plane. Now that we have the 3D position of every person in the scene, the dynamic global risk in Eq.~\ref{eq:D} can be calculated and given as output along with other information that we summarize in the following subsection.

\subsection{System Output}
\label{sec:output}
A convenient graphical interface highlights all the main results of the analysis of our Inter-Homines system, allowing to evaluate at a glance the crowding conditions in the monitored area (see Fig.~\ref{fig:gui}). This interface is made with Qt to guarantee compatibility with all the main operating systems.

\paragraph{Anonymized Frame} It shows real-time the bounding box detections superimposed to the input RGB frame. The system is privacy compliant and all the faces are obscured. Colored segments connect people who are at an estimated distance lower than a defined upper threshold distance (typically 3 m). The color indicates the extent of the infraction, going from a dark red for the most serious infraction to yellow for the minor ones.

\paragraph{People Counter} At the top right of the frame we also display the number of detected people updated in real time. This number is an average computed in a window of $W$ frames to account for miss detections and false positives.

\paragraph{Dynamic Risk and Occupation Maps} In the right part of the interface two bird eye views of the walking area are updated real-time. The Dynamic Risk map shows a snapshot of the current situation of the area. The Occupation map, instead, displays the overall aggregated risk and it is computed by averaging the Dynamic Risk maps of the whole day. It is used to identify areas with a larger risk for statistical and predictive purposes. Note that people outside of the walking area are completely ignored and do not affect the statistics.

\paragraph{Estimated Warning Level} In the lower part of the window a bar represents the total estimated risk in the monitored area and it is computed using Eq.~\ref{eq:D}. The application provides the possibility to send an alarm signal (example: send an email / audio notification) if certain thresholds on the number of people or on the risk level are exceeded. The thresholds and the notification methods of their exceeding will be defined according to the needs of the context in which the system will operate.

\paragraph{Weekly Report} Since we want to give insightful statistics to help with the prevention of the infection, our system periodically produce a report. The report contains statistics about number of people, risk level, number of infractions and occupation maps aggregated by hours and days. To this end we utilize a non-relational database to store timestamp and position of each person captured by our system.

\begin{figure*}[t]
\begin{center}
\includegraphics[width=0.32\textwidth]{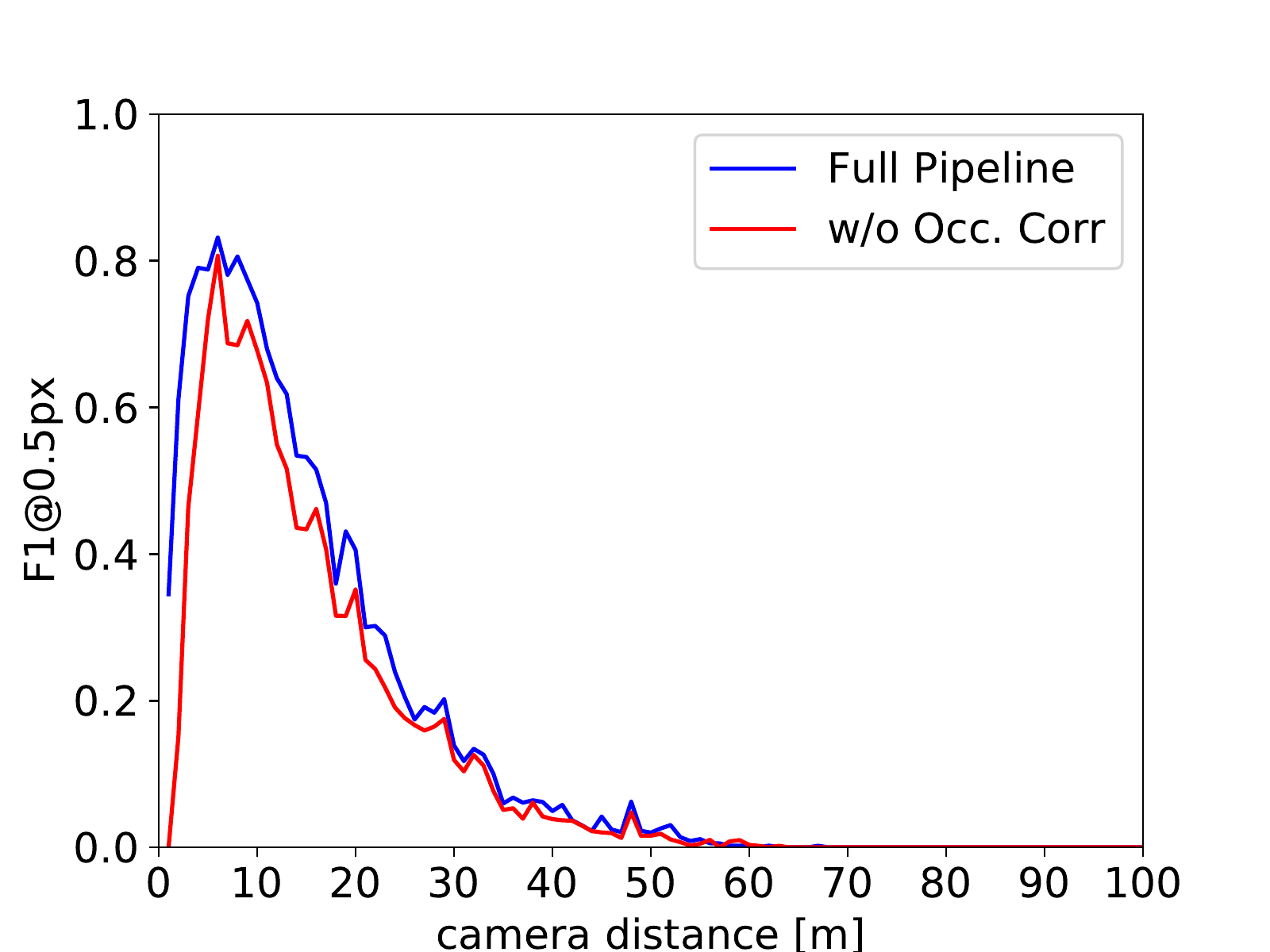}
\includegraphics[width=0.32\textwidth]{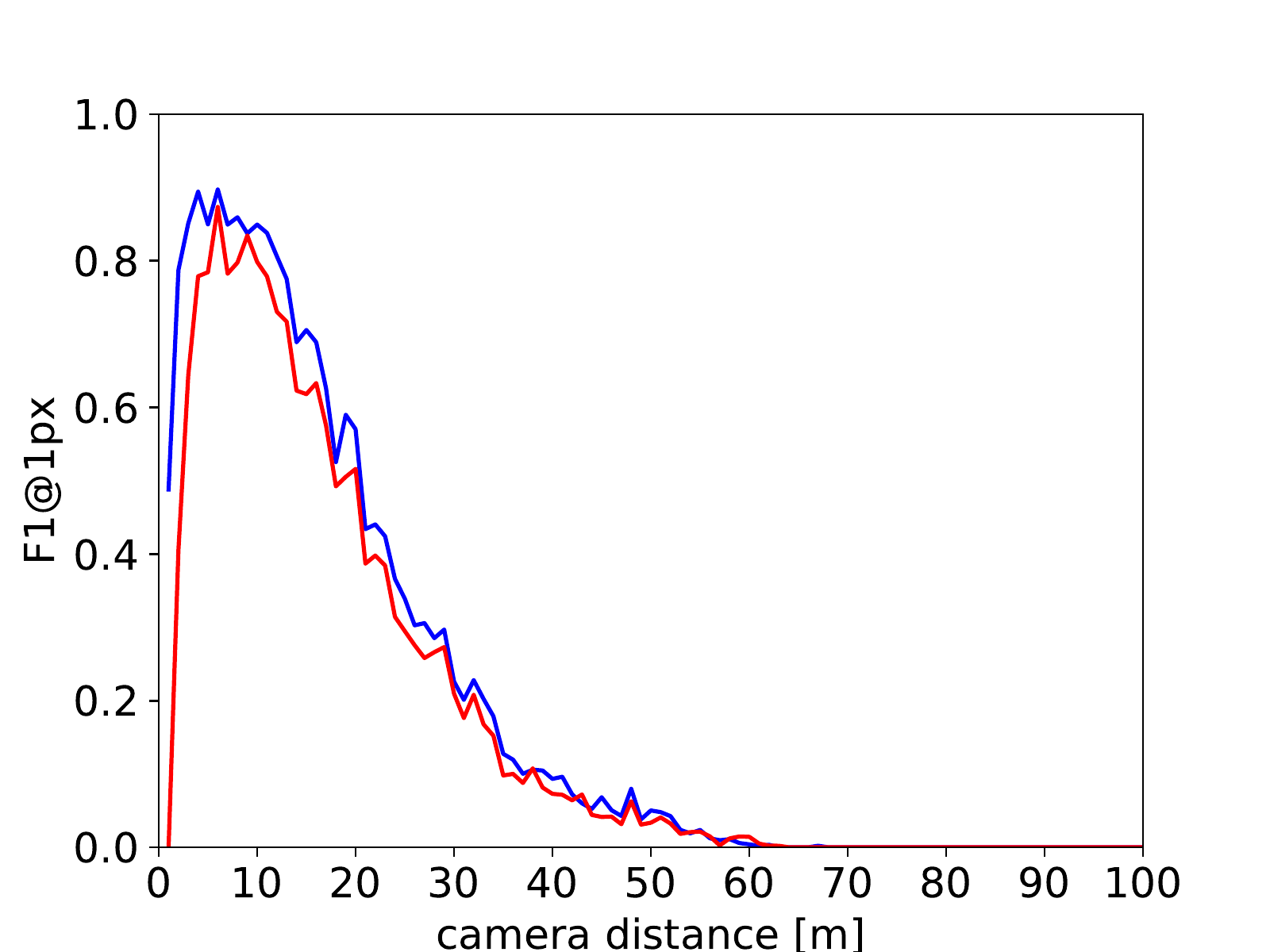}
\includegraphics[width=0.32\textwidth]{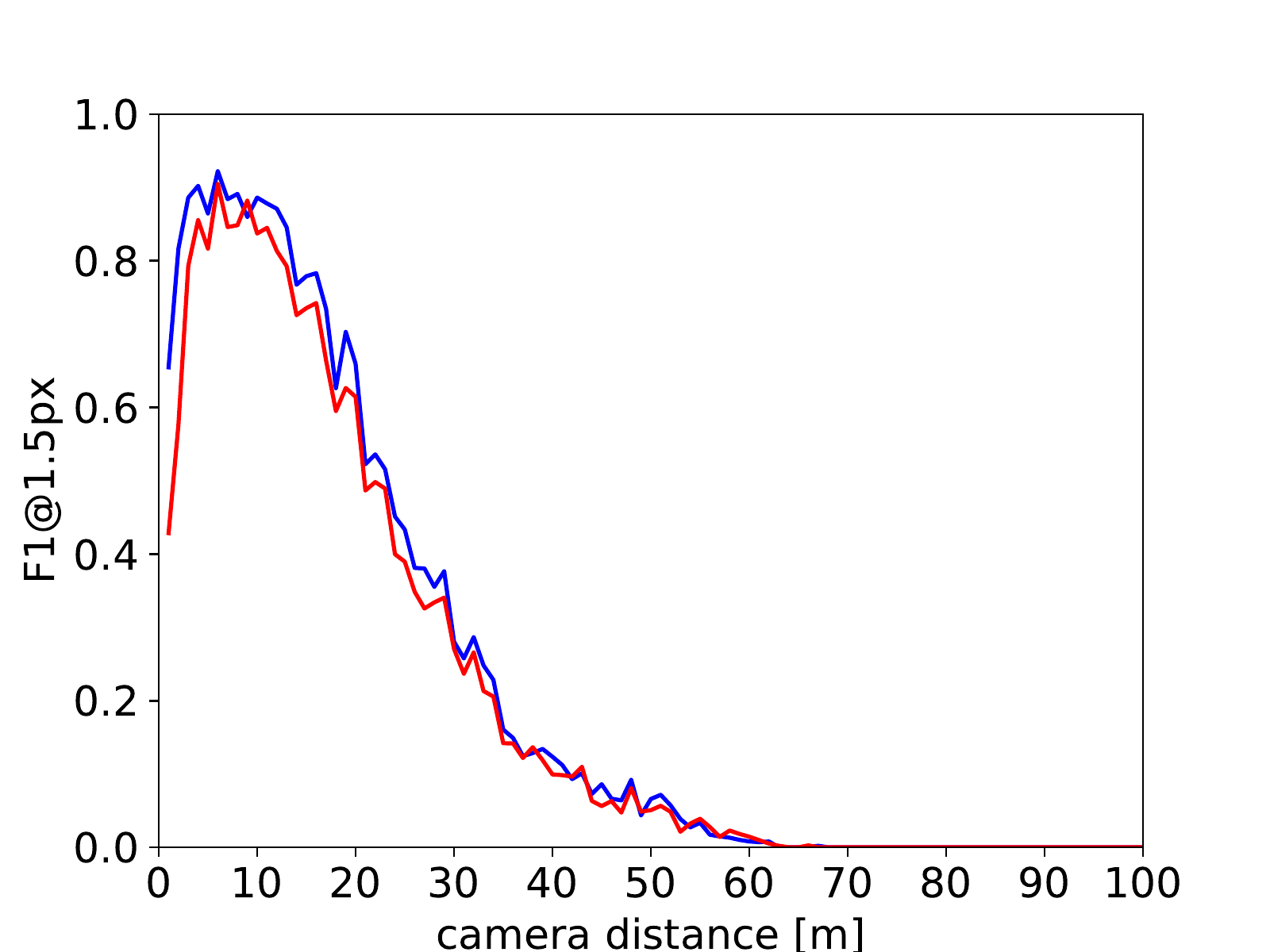}
\\[1px]
\includegraphics[width=0.32\textwidth]{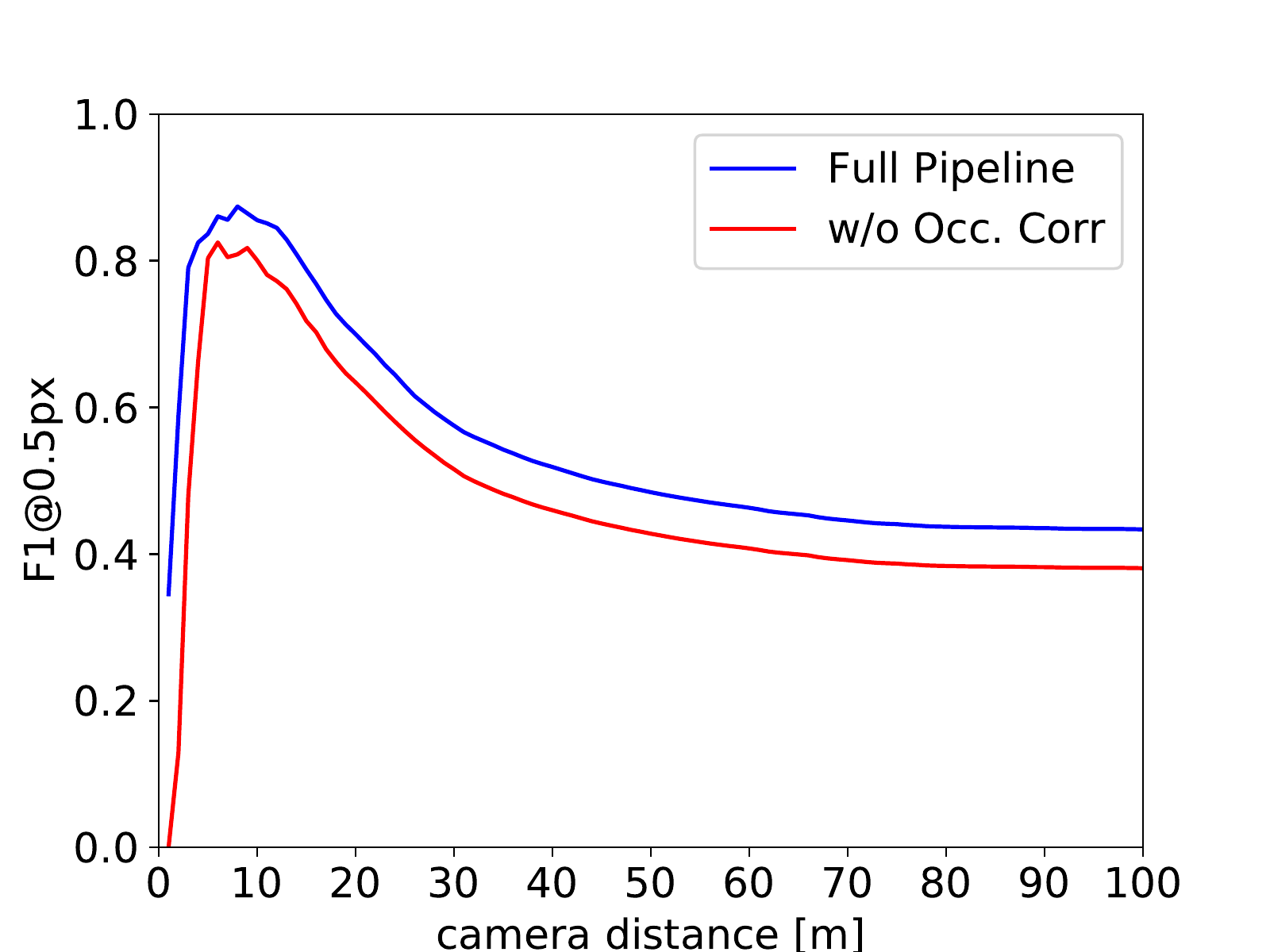}
\includegraphics[width=0.32\textwidth]{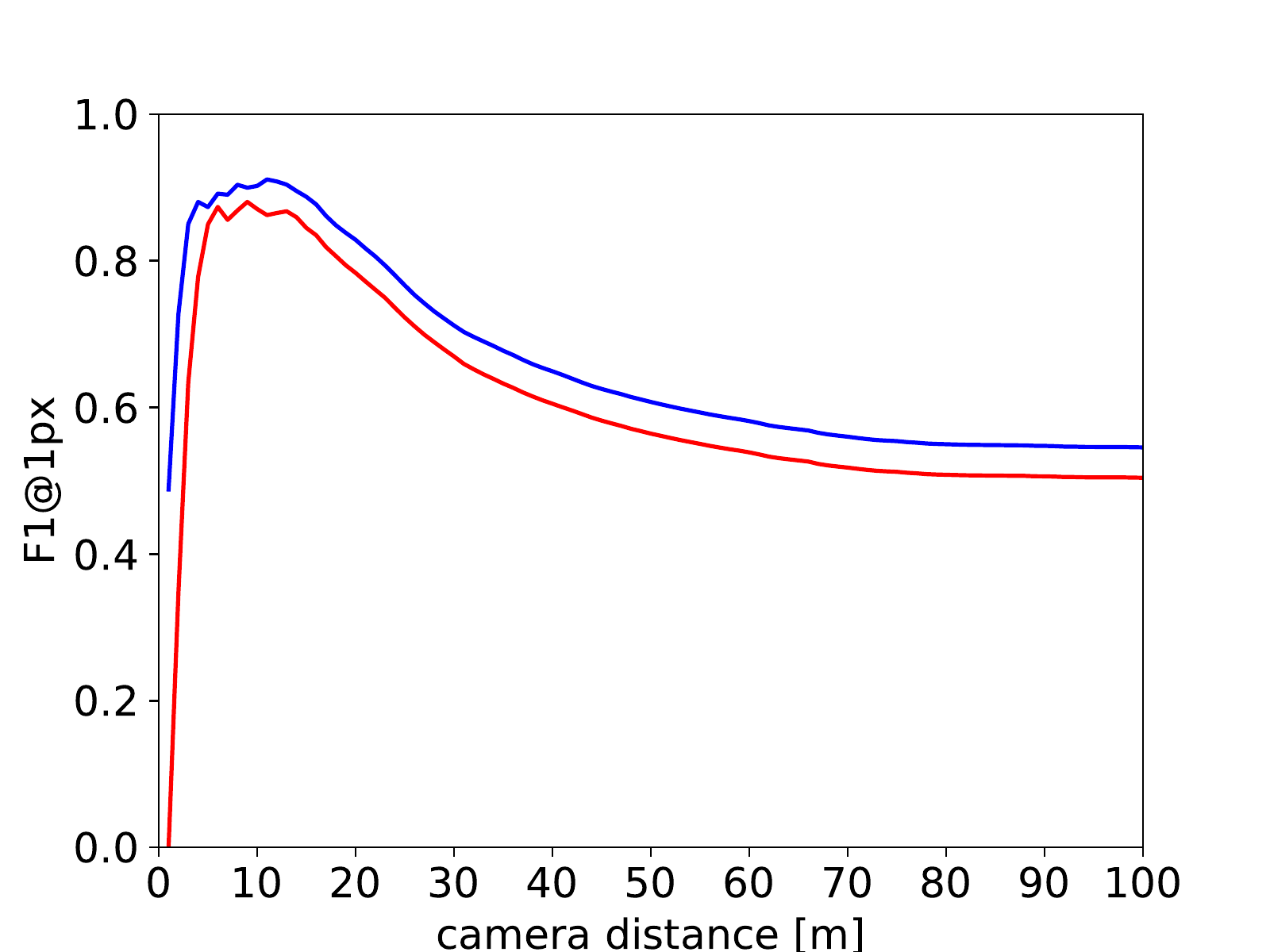}
\includegraphics[width=0.32\textwidth]{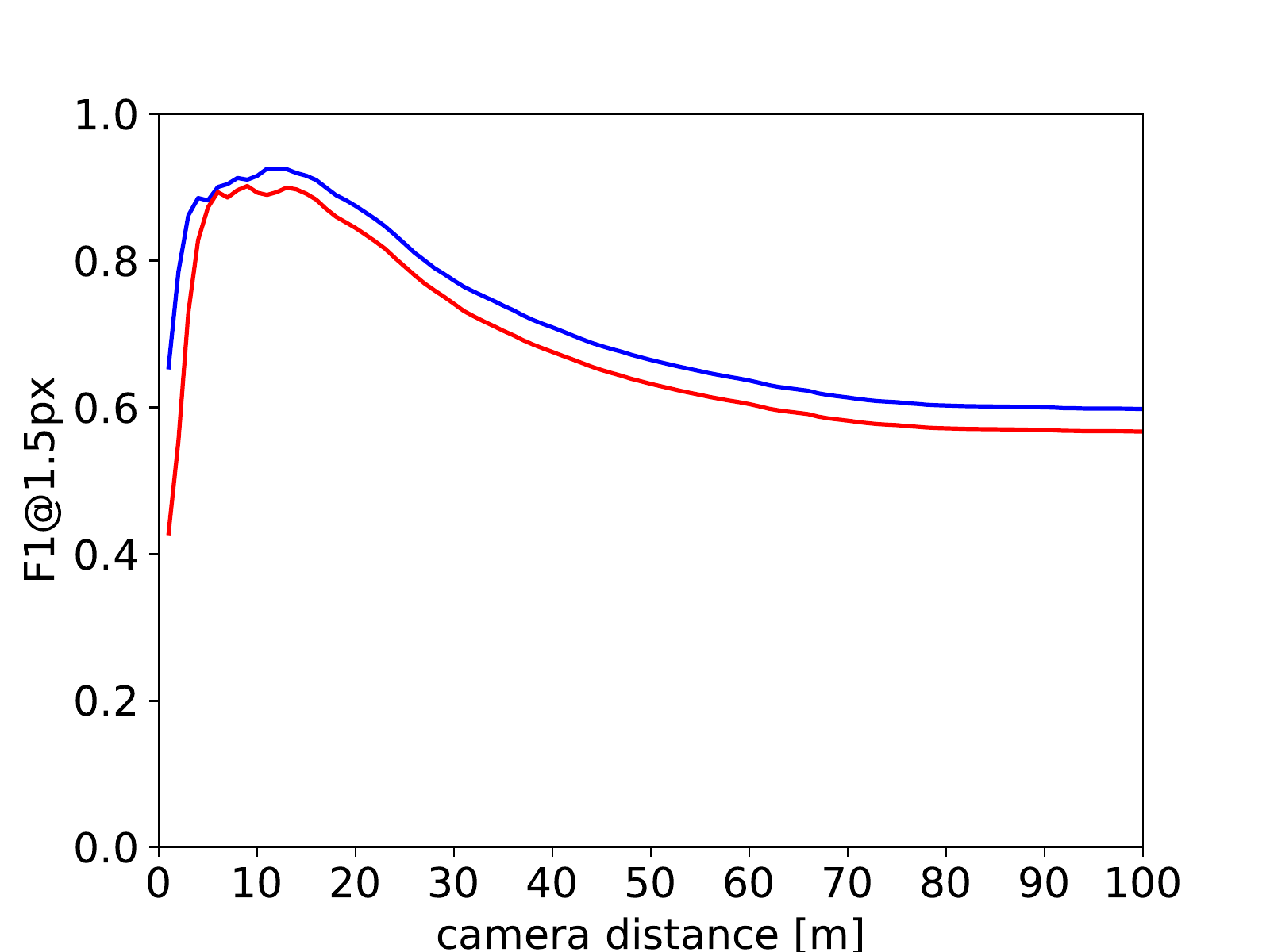}
\end{center}
\caption{F1 score vs. camera distance at different thresholds (first row) and F1 score vs. max camera distance at different thresholds (second row).}
\label{fig:graphs}
\end{figure*}

\section{System Validation}
In order to validate the effectiveness of our system, we performed a series of experiments leveraging JTA \cite{fabbri2018learning}. JTA is a massive dataset for pedestrian pose estimation and tracking in urban scenarios created exploiting the highly photorealistic video game \textit{Grand Theft Auto V}. The videos feature a vast number of different people appearances, in several urban scenarios at varying illumination conditions and viewpoints. Each clip comes with a precise annotation of visible and occluded body parts, people tracking with 2D coordinates in image plane and 3D coordinates in camera space. JTA overcomes all the limitation of existing datasets in terms of number of entities and available annotations. Each video contains a number of people ranging between 0 and 60 with an average of more than 21 people, totaling almost 10M annotated body poses over 460,800 densely annotated frames. The distance from the camera ranges between 0.1 and 100 meters, resulting in pedestrian heights between 20 and 1100 pixels. JTA is composed by a set of 512 Full HD videos, 30 seconds long, recorded at 30 fps. 

As shown in Fig.~\ref{fig:JTA}, despite being a synthetic dataset, JTA features highly challenging and complex situations, peculiar of surveillance scenarios, where people are often dominated by severe body part occlusions and truncations. We believe this dataset is the perfect choice to validate a system that targets global safety. 

Since we can not perform the system calibration procedure on an already recorded dataset, i.e. we can not physically place the markers at the center of the scene, we designed a simple heuristic to directly recover the nine points pairs using the dataset annotations. With the assumption that every foot of each person lies on the same plane, for each JTA sequence, we linearly regressed the ground plane utilizing the 3D coordinates of every foot in every frame of that sequence. By recovering a unit normal vector of the plane and two orthonormal vectors lying on the plane we were able to find the orthonormal base of the new space that allowed us to move each 3D coordinate into a space where each foot has the same $y$ coordinate (according to the standard camera system). Now, since each foot coordinate has the same $y$, we can get rid of it and considering the new $(x,z)$ coordinates as ground coordinates. As we are interested in nine points pairs of target and source planes, we utilized a K-Means implementation to find nine foot cluster centers. Utilizing a clustering method ensures that the nine points are far from each other. Once recovered the foot cluster centers, we remapped those coordinates into the original standard camera space and projected them into the image plane using the pinhole camera model. The 2D projected coordinates and the 2D foot clusters now form the nine points pairs needed to calculate the homography matrix.

Experiments are conducted on every 10\textsuperscript{th} frame of a subset of the JTA test set where we carefully removed the sequences that contain camera motion and people at different heights, e.g. people going up the stairs, as our method assumes static camera and flat ground plane. Tab.~\ref{tab:cp_exp_jta} shows the precision, recall and F1 obtained using different thresholds and considering different camera distance ranges. As the range increase, we observe a decrease in performances, due to the fact that small people are hardly detected and homography transformation becomes less reliable. Since we are interested in evaluating the impact of that occlusions have in the performance of our system, we reported the results with and without the occlusion correction module. As can be shown, the correction is always beneficial, especially when people are close to the camera. 

To better understand how performance degrades as distance increases, in Fig.~\ref{fig:graphs}, first row, we plotted the F1 score at different thresholds w.r.t. the camera distance. It is interesting to note that performance worsens when people are too close to the camera. In Fig.~\ref{fig:graphs}, second row, we plotted the same quantity but, this time, the F1 score is calculated considering all the people with distance less than the camera distance, and not equal to the camera distance.

\section{Conclusions}
In this work we proposed a simple and effective system that deals with the COVID-19 emergency by providing a social distancing tool that can prevent the spread of the infection. We validated it using a highly challenging benchmark, obtaining a lower bound on the performance of the method. We believe that our system can be a practical solution to an important problem, hoping to see areas less crowded than the JTA dataset in the near future.

\section*{Acknowledgments}
The work is supported by the Italian MUR, Ministry of Universities and Research, under the project PRIN 2019-2021 ``PREVUE Prediction of Events in Urban Environment'' project and partially supported by EU European Regional Development Funds for Regione Emilia Romagna under the special project ``Inter-Homines'' 2020 among the 16 research and innovation projects for the development of solutions aimed at contrasting the epidemic of COVID-19.

\bibliographystyle{IEEEtran}
\bibliography{egbib}

\begin{thebibliography}{10}
\providecommand{\url}[1]{#1}
\csname url@samestyle\endcsname
\providecommand{\newblock}{\relax}
\providecommand{\bibinfo}[2]{#2}
\providecommand{\BIBentrySTDinterwordspacing}{\spaceskip=0pt\relax}
\providecommand{\BIBentryALTinterwordstretchfactor}{4}
\providecommand{\BIBentryALTinterwordspacing}{\spaceskip=\fontdimen2\font plus
\BIBentryALTinterwordstretchfactor\fontdimen3\font minus
  \fontdimen4\font\relax}
\providecommand{\BIBforeignlanguage}[2]{{%
\expandafter\ifx\csname l@#1\endcsname\relax
\typeout{** WARNING: IEEEtran.bst: No hyphenation pattern has been}%
\typeout{** loaded for the language `#1'. Using the pattern for}%
\typeout{** the default language instead.}%
\else
\language=\csname l@#1\endcsname
\fi
#2}}
\providecommand{\BIBdecl}{\relax}
\BIBdecl

\bibitem{asadi2020coronavirus}
S.~Asadi, N.~Bouvier, A.~S. Wexler, and W.~D. Ristenpart, ``The coronavirus
  pandemic and aerosols: Does covid-19 transmit via expiratory particles?''
  2020.

\bibitem{chen2020time}
Y.-C. Chen, P.-E. Lu, C.-S. Chang, and T.-H. Liu, ``A time-dependent sir model
  for covid-19 with undetectable infected persons,'' \emph{arXiv:2003.00122},
  2020.

\bibitem{fabbri2018learning}
M.~Fabbri, F.~Lanzi, S.~Calderara, A.~Palazzi, R.~Vezzani, and R.~Cucchiara,
  ``Learning to detect and track visible and occluded body joints in a virtual
  world,'' in \emph{Proceedings of the European Conference on Computer Vision
  (ECCV)}, 2018.

\bibitem{fabbri2020compressed}
M.~Fabbri, F.~Lanzi, S.~Calderara, S.~Alletto, and R.~Cucchiara, ``Compressed
  volumetric heatmaps for multi-person 3d pose estimation,'' in
  \emph{Proceedings of the IEEE/CVF Conference on Computer Vision and Pattern
  Recognition}, 2020.

\bibitem{girshick2014rich}
R.~Girshick, J.~Donahue, T.~Darrell, and J.~Malik, ``Rich feature hierarchies
  for accurate object detection and semantic segmentation,'' in
  \emph{Proceedings of the IEEE conference on computer vision and pattern
  recognition}, 2014.

\bibitem{xiang2017subcategory}
Y.~Xiang, W.~Choi, Y.~Lin, and S.~Savarese, ``Subcategory-aware convolutional
  neural networks for object proposals and detection,'' in \emph{IEEE winter
  conference on applications of computer vision (WACV)}, 2017.

\bibitem{girshick2015fast}
R.~Girshick, ``Fast r-cnn,'' in \emph{Proceedings of the IEEE international
  conference on computer vision}, 2015.

\bibitem{ren2015faster}
S.~Ren, K.~He, R.~Girshick, and J.~Sun, ``Faster r-cnn: Towards real-time
  object detection with region proposal networks,'' in \emph{Advances in neural
  information processing systems}, 2015.

\bibitem{redmon2017yolo9000}
J.~Redmon and A.~Farhadi, ``Yolo9000: better, faster, stronger,'' in
  \emph{Proceedings of the IEEE conference on computer vision and pattern
  recognition}, 2017.

\bibitem{redmon2018yolov3}
------, ``Yolov3: An incremental improvement,'' \emph{arXiv:1804.02767}, 2018.

\bibitem{liu2016ssd}
W.~Liu, D.~Anguelov, D.~Erhan, C.~Szegedy, S.~Reed, C.-Y. Fu, and A.~C. Berg,
  ``Ssd: Single shot multibox detector,'' in \emph{European conference on
  computer vision}, 2016.

\bibitem{lin2017focal}
T.-Y. Lin, P.~Goyal, R.~Girshick, K.~He, and P.~Doll{\'a}r, ``Focal loss for
  dense object detection,'' in \emph{Proceedings of the IEEE international
  conference on computer vision}, 2017.

\bibitem{zhou2019objects}
X.~Zhou, D.~Wang, and P.~Kr{\"a}henb{\"u}hl, ``Objects as points,''
  \emph{arXiv:1904.07850}, 2019.

\bibitem{bodla2017soft}
N.~Bodla, B.~Singh, R.~Chellappa, and L.~S. Davis, ``Soft-nms--improving object
  detection with one line of code,'' in \emph{Proceedings of the IEEE
  international conference on computer vision}, 2017.

\bibitem{cao2017realtime}
Z.~Cao, T.~Simon, S.-E. Wei, and Y.~Sheikh, ``Realtime multi-person 2d pose
  estimation using part affinity fields,'' in \emph{Proceedings of the IEEE
  conference on computer vision and pattern recognition}, 2017.

\bibitem{he2016deep}
K.~He, X.~Zhang, S.~Ren, and J.~Sun, ``Deep residual learning for image
  recognition,'' in \emph{Proceedings of the IEEE conference on computer vision
  and pattern recognition}, 2016.

\bibitem{he2017mask}
K.~He, G.~Gkioxari, P.~Doll{\'a}r, and R.~Girshick, ``Mask r-cnn,'' in
  \emph{IEEE international conference on computer vision}, 2017.

\bibitem{law2018cornernet}
H.~Law and J.~Deng, ``Cornernet: Detecting objects as paired keypoints,'' in
  \emph{Proceedings of the European Conference on Computer Vision}, 2018.

\bibitem{zhou2019bottom}
X.~Zhou, J.~Zhuo, and P.~Krahenbuhl, ``Bottom-up object detection by grouping
  extreme and center points,'' in \emph{Proceedings of the IEEE Conference on
  Computer Vision and Pattern Recognition}, 2019.

\bibitem{newell2016stacked}
A.~Newell, K.~Yang, and J.~Deng, ``Stacked hourglass networks for human pose
  estimation,'' in \emph{European conference on computer vision}, 2016.

\bibitem{insafutdinov2016deepercut}
E.~Insafutdinov, L.~Pishchulin, B.~Andres, M.~Andriluka, and B.~Schiele,
  ``Deepercut: A deeper, stronger, and faster multi-person pose estimation
  model,'' in \emph{European Conference on Computer Vision}, 2016.

\bibitem{kundu20183d}
A.~Kundu, Y.~Li, and J.~M. Rehg, ``3d-rcnn: Instance-level 3d object
  reconstruction via render-and-compare,'' in \emph{Proceedings of the IEEE
  conference on computer vision and pattern recognition}, 2018.

\bibitem{chabot2017deep}
F.~Chabot, M.~Chaouch, J.~Rabarisoa, C.~Teuliere, and T.~Chateau, ``Deep manta:
  A coarse-to-fine many-task network for joint 2d and 3d vehicle analysis from
  monocular image,'' in \emph{Proceedings of the IEEE conference on computer
  vision and pattern recognition}, 2017.

\bibitem{mousavian20173d}
A.~Mousavian, D.~Anguelov, J.~Flynn, and J.~Kosecka, ``3d bounding box
  estimation using deep learning and geometry,'' in \emph{Proceedings of the
  IEEE Conference on Computer Vision and Pattern Recognition}, 2017.

\bibitem{kermack1927contribution}
W.~O. Kermack and A.~G. McKendrick, ``A contribution to the mathematical theory
  of epidemics,'' \emph{The royal society of london. Series A, Containing
  papers of a mathematical and physical character}, 1927.

\bibitem{yu2018deep}
F.~Yu, D.~Wang, E.~Shelhamer, and T.~Darrell, ``Deep layer aggregation,'' in
  \emph{Proceedings of the IEEE conference on computer vision and pattern
  recognition}, 2018.

\bibitem{lin2014microsoft}
T.-Y. Lin, M.~Maire, S.~Belongie, J.~Hays, P.~Perona, D.~Ramanan,
  P.~Doll{\'a}r, and C.~L. Zitnick, ``Microsoft coco: Common objects in
  context,'' in \emph{European conference on computer vision}, 2014.

\bibitem{hartley2003multiple}
R.~Hartley and A.~Zisserman, \emph{Multiple view geometry in computer
  vision}.\hskip 1em plus 0.5em minus 0.4em\relax Cambridge university press,
  2003.

\end{thebibliography}

\begin{IEEEbiography}
[{\includegraphics[width=1in,height=1.25in,clip,keepaspectratio]{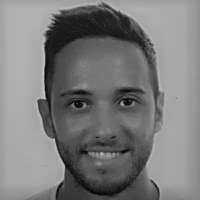}}]{Matteo Fabbri}
is currently a Ph.D. student at the International Doctorate School in ICT of the University of Modena e Reggio Emilia. He works under the supervision of Prof. Rita Cucchiara and Prof. Simone Calderara, on computer vision and deep learning for people behavior understanding. He worked 10 months at Panasonic Silicon Valley Lab as a Deep Learning Engineer. His research interests include generative models, pose estimation and multiple object tracking.
\end{IEEEbiography}
\begin{IEEEbiography}
[{\includegraphics[width=1in,height=1.25in,clip,keepaspectratio]{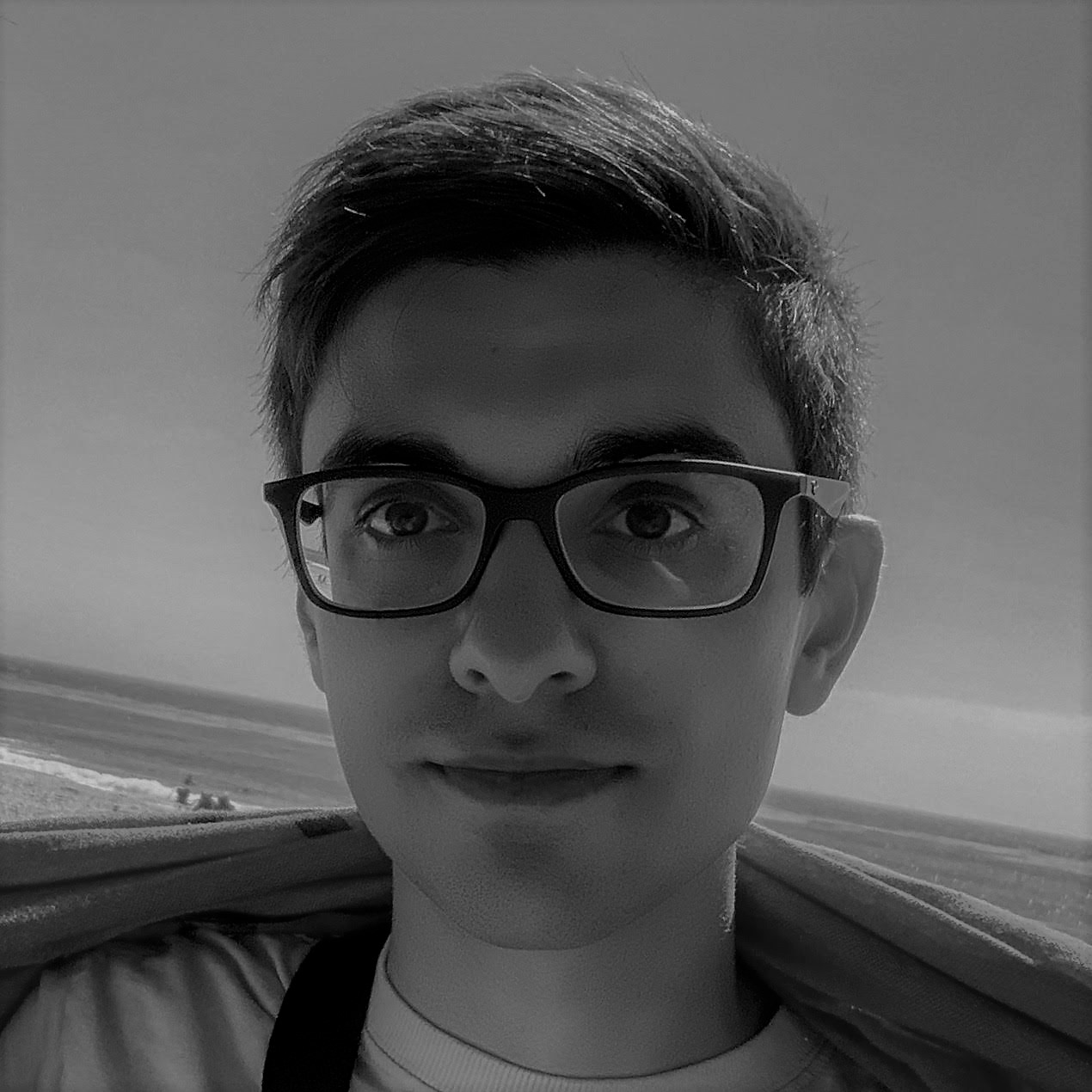}}]{Fabio Lanzi}
is currently a research fellow at the Artificial Intelligence Research and Innovation Center (AIRI) promoted by the "Enzo Ferrari" Department of Engineering and by the "Marco Biagi" Department of Economics of the University of Modena and Reggio Emilia. Since 2017, he works under the supervision of Prof. Rita Cucchiara and Prof. Simone Calderara, and he has he took part in a series of industrial research projects mainly concerning human pose estimation, tracking and action recognition.
\end{IEEEbiography}
\begin{IEEEbiography}
[{\includegraphics[width=1in,height=1.25in,clip,keepaspectratio]{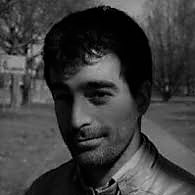}}]{Riccardo Gasparini}
is currently a research fellow at AImagelab, a research laboratory of the Department of Engineering "Enzo Ferrari" at the University of Modena and Reggio Emilia, Italy. He works under the supervision of Prof. Rita Cucchiara on various topics such People Detection, Tracking, People Recognition, Video Surveillance, Anomaly Detection, Video Analysis, Egocentric vision and Embedded sensors.
\end{IEEEbiography}
\begin{IEEEbiography}
[{\includegraphics[width=1in,height=1.25in,clip,keepaspectratio]{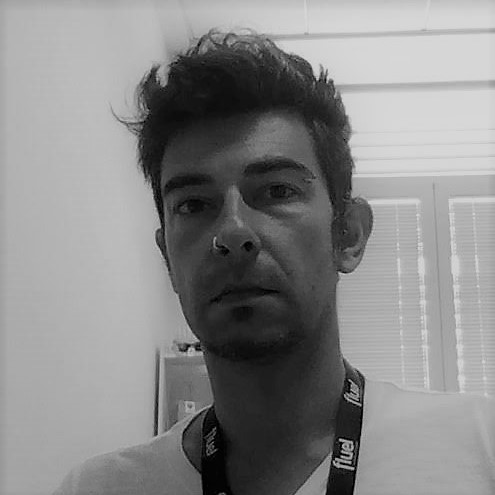}}]{Simome Calderara}
received a computer engineering masters degree in 2005 and the Ph.D. degree in 2009 from the University of Modena and Reggio Emilia, where he is currently an assistant professor within the AImageLab group. His current research interests include computer vision and machine learning applied to human behavior analysis, visual tracking in crowded scenarios, and time series analysis for forensic applications. He is a member of the IEEE.
\end{IEEEbiography}
\begin{IEEEbiography}
[{\includegraphics[width=1in,height=1.25in,clip,keepaspectratio]{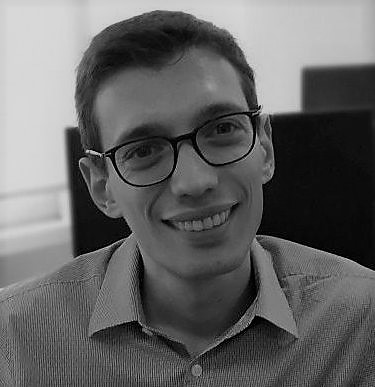}}]{Lorenzo Baraldi}
received the Ph.D. degree (cum laude) in information and communication technologies from the Università degli studi di Modena e Reggio Emilia, Italy, in 2018. He was a Research Intern with Facebook AI Research (FAIR) in 2017. He is currently an Assistant Professor with the Dipartimento di Ingegneria “Enzo Ferrari”, Università degli Studi di Modena e Reggio Emilia. He has authored or coauthored over 50 publications in scientific journals and international conference proceedings. His research interests include image processing, video understanding, deep learning and multimedia.
\end{IEEEbiography}
\begin{IEEEbiography}
[{\includegraphics[width=1in,height=1.25in,clip,keepaspectratio]{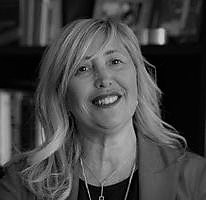}}]{Rita Cucchiara}
received the master’s degree in Electronic Engineering and the Ph.D. degree in Computer Engineering from the University of Bologna, Italy, in 1989 and 1992, respectively. Since 2005, she is a full professor at the University of Modena and Reggio Emilia, Italy, where she heads the AImageLab group and is Director of the CINI AIIS Lab. She published more than 300 papers on pattern recognition computer vision and multimedia, and in particular in human analysis, HBU, and egocentric-vision. The research carried out spans on different application fields, such as video surveillance, automotive and multimedia big data annotation. Currently, she is AE of IEEE Transactions on Multimedia and serves in the Governing Board of IAPR and in the Advisory Board of the CVF.
\end{IEEEbiography}

\end{document}